\documentclass[runningheads]{llncs}
\usepackage{lineno,hyperref}
\usepackage{xcolor}
\usepackage{booktabs}
\usepackage{hyperref}
\usepackage{colortbl}
\usepackage{multicol}
\usepackage{mathtools}
\usepackage{amsmath}
\usepackage{amssymb}
\usepackage{pifont}
\usepackage{multirow}
\usepackage{makecell}
\usepackage{subcaption}
\usepackage{manyfoot}%
\usepackage{booktabs}%
\usepackage{listings}%
\usepackage{enumitem}
\usepackage{makecell}
\usepackage{booktabs}
\usepackage{subcaption}
\usepackage{bbding}
\usepackage{bbm}
\usepackage{graphicx}
\usepackage{tabularx}
\usepackage{multirow}
\usepackage[T1]{fontenc}
\usepackage{bbding}
\usepackage[marginal]{footmisc}

\usepackage{colortbl}
\usepackage{tabularx}
\usepackage{multirow}

\usepackage{graphicx}
\begin{document}
%
\title{KA$^2$ER: Knowledge Adaptive Amalgamation of ExpeRts for Medical Images Segmentation}
%
%
\author{Shangde Gao\inst{1,2}
\and Yichao Fu\inst{1}
\and Ke Liu\inst{1}
\and Hongxia Xu\Envelope\inst{2,3}
\and Jian Wu \inst{4,5}}

\authorrunning{Shangde Gao, et al.}
\institute{College of Computer Science and Technology, Zhejiang University, Hangzhou, China
\and Liangzhu Laboratory, 
Zhejiang University, Hangzhou, China
\and WeDoctor Holdings Limited, Hangzhou, China
\and 
State Key Laboratory of Transvascular Implantation Devices of The Second Affiliated Hospital, Zhejiang University School of Medicine, Hangzhou, China
\and 
School of Public Health Zhejiang University, Hangzhou, China\\
\email{\{gaosde,fuyichao, lk2017, Einstein, wujian2000\}@zju.edu.cn}}
\maketitle
\begin{abstract}
Recently, many foundation models for medical image analysis such as MedSAM, SwinUNETR have been released and proven to be useful in multiple tasks.
However, considering the inherent heterogeneity and inhomogeneity of real-world medical data, directly applying these models to specific medical image segmentation tasks often leads to negative domain shift effects, which can severely weaken the model's segmentation capabilities. 
To this end, we propose an adaptive amalgamation knowledge framework that aims to train a versatile foundation model to handle the joint goals of multiple expert models, each specialized for a distinct task. 
Specifically, we first train an nnUNet-based expert model for each task, and reuse the pre-trained SwinUNTER as the target foundation model.
Then, the input data for all challenging tasks are encoded in the foundation model and the expert models, respectively, and their backbone features are jointly projected into the adaptive amalgamation layer.
Within the hidden layer, the hierarchical attention mechanisms are designed to achieve adaptive merging of the target model to the hidden layer feature knowledge of all experts, which significantly reduces the domain shift arising from the inter-task differences. 
Finally, the gold amalgamated features and the prompt features are fed into the mask decoder to obtain the segmentation results.
Extensive experiments conducted in these challenging tasks demonstrate the effectiveness and adaptability of our foundation model for real-world medical image segmentation.

\end{abstract}
\section{Introduction}

Medical image segmentation aims to classify anatomical structures of various organs using medical imaging techniques~\cite{zhuang2013challenges,boyle2019computationally}. 
The rapid evolution of deep learning methodologies has significantly enhanced its efficacy in medical image analysis, 
finding extensive applications in clinical early cancer detection~\cite{cao2023large,xu2023group} and disease diagnosis~\cite{Zhuang2019evaluation,li2022Survey}. 
The availability of large-scale, diverse training datasets and robust computational capabilities has recently facilitated the development of foundation models 
like SAM~\cite{kirillov2023segment}, MedSAM~\cite{ma2024segment}, and SwinUNETR~\cite{tang2022self}, enabling the segmentation of various image objects and scenes, 
with capabilities extending to zero-shot generalization across novel domains. 
However, their performance on real-world data remains inadequately explored. 
Specifically, foundation models are typically trained on standard and large anatomical targets such as the liver and lungs, 
whereas real-world datasets often originate from various centers and encompass varied imaging sequences and modalities~\cite{zhuang2019multivariate}. 
Furthermore, the challenge persists in directly applying these foundation models to specific segmentation tasks 
involving small and irregularly shaped regions of interest (ROIs), such as lesions or scars.

For a specific dataset in a segmentation task, machine learning methods typically achieve good performance under the assumption that the training and test data are independently and identically distributed, and that the expert model is trained using supervised learning strategies~\cite{gao2022joint,GAO2023BayeSeg}. However, due to the inherent characteristics of medical imaging, even within the same segmentation task, medical images from different sequences, modalities, and sites often exhibit varying intensities, leading to distributional differences. As a result, expert models trained on one dataset often struggle to generalize to unseen datasets, necessitating the costly process of training new models from scratch. Consequently, developing effective and efficient transfer learning methods that enable foundation and expert models to complement each other’s strengths is of significant value for real-world medical image segmentation.

To fully utilize the foundation models and experts for medical image segmentation, we propose a framework, \textbf{K}nowledge \textbf{A}daptive \textbf{A}malgamation of \textbf{E}xpe\textbf{R}ts (\textbf{KA$^2$ER}), maintaining the generalization capabilities of foundation models while incorporating domain-specific knowledge from expert models. 

Specifically, we first train task-specific expert models on each sub-task, which is split by the modality and sequence of data. 
Then we adopt SwinUNETR as the foundation student model, and all data are jointly fed into both expert and foundation models to extract backbone network features, with expert  model parameters 
frozen during this process. 
To improve the training efficiency, we design a mapping layer to align task-specific expert models with segmentation prompts, filtering only the task-specific expert model features that are relevant to the data in the current mini-batch, rather than feeding all expert features into the adaptive amalgamation layer.
Subsequently, the filtered task-specific expert features, task-agnostic expert features, and student model features are processed in the amalgamation layer, where the student model is amalgamated with these features using
hierarchical attention module.
Finally, the amalgamated features enter the student model decoder, and the output from the last layer of the decoder is concatenated with the pre-embedded prompts into a lightweight mask decoder, producing the final binary segmentation result.

The main contributions of this work are summarized as follows:
\begin{itemize}
\item 
We present a new framework for knowledge adaptive amalgamation of experts, i.e. KA$^2$ER, which preserves the generalizability of the foundation model while enabling it to adaptively merge the discriminative knowledge of pre-trained experts for real-world medical image segmentation tasks.
\item 
The adaptive amalgamation module enables the foundation model to learn discriminative feature knowledge from task-specific and task-agnostic experts in the hidden feature space.
\item 
Extensive experiments are conducted on four datasets from the MICAAI`2024 CARE challenge and empirical results demonstrate the effectiveness of our KA$^2$ER for real-world medical image segmentation tasks.
\end{itemize}

To the best of our knowledge, this is the first work to transfer knowledge amalgamation from natural to medical images and amalgamate foundation and multi-expert models for medical image segmentation tasks.
\section{Related Works}
\subsection{Medical Image Segmentation}
CNN-based and Transformer-based foundational models have greatly advanced medical image segmentation. 
U-Net\cite{ronneberger2015u}, a prominent CNN-based model, employs a symmetric encoder-decoder architecture with skip connections, demonstrating strong performance across various medical segmentation tasks~\cite{Wu2023SemiSL,li2021AtrialGeneral,li2022AtrialJSQnet,liu2024merit,Zhuang2019evaluation}. 
In~\cite{qiu2023myops}, the authors introduced MyoPS-Net, an end-to-end deep neural network designed to integrate five-sequence cardiac magnetic resonance (CMR) images for the segmentation of myocardial pathology. 
Gao et al.~\cite{gao2022joint,GAO2023BayeSeg} proposed BayeSeg, a novel framework that uses Bayesian modeling to refine medical image segmentation. 
Recent advancements in foundational models, driven by large-scale pretraining data, have significantly improved segmentation tasks. SAM~\cite{kirillov2023segment}, as a versatile segmentation model, excels in natural image segmentation by supporting diverse tasks through prompt-based inputs. 
MedSAM~\cite{ma2024segment}, extends this adaptability to medical imaging by pretraining on 2D medical data.
Swin UNETR~\cite{tang2022self} further advances performance by integrating Swin Transformer modules and sliding window inference. SegVol~\cite{du2023segvol}, otherwise, focuses on modeling 3D volumetric medical data, 
making it a key tool for large-scale volumetric dataset analysis. 
Despite these advancements, challenges persist in medical image segmentation, such as variability in organ shapes, motion artifacts, contrast-to-noise ratio issues, domain shifts in multi-center datasets, and modality discrepancies.
\subsection{Transfer Learning}
Transfer learning aims to enhance performance in the target domain by leveraging knowledge from related source domains~\cite{HintonVD15}. 
Knowledge distillation (KD), as a prominent strategy, utilizes comprehensive teacher models from the source domain to train more compact student models in the target domain~\cite{HintonVD15}, includes methods such as intermediate layer feature alignment~\cite{yang2023knowledge}), and logits-based distillation~\cite{xu2024survey}). 
Knowledge amalgamation (KA) represents a more comprehensive approach by developing a compact student model that integrates knowledge from multiple task-specific teacher models, thereby enhancing the robustness and generalization performance of the model in complex domains. 
Compared to KD, this approach can effectively mitigate the adverse effects of distribution shift and has been applied to challenging tasks 
~\cite{gao2023contrastive,gao2024collaborative,zhang2023knowledge}. 
Concretely, Shen et al.~\cite{shen2019amalgamating} align the probabilities of the student with the combined teachers, while most approaches~\cite{gao2023contrastive,gao2024collaborative} minimize the distance by aligning features of the student and teachers in a shared latent space. 
However, previous works predominantly focused on natural image classification and segmentation, where training data are kept independently and identically distributed (i.e., IID), leaving the complex scenario of open-world medical image segmentation largely unexplored. 
\section{Methodology} 
\subsection{Preliminary of Knowledge Amalgamation
}
Knowledge amalgamation (KA) aims to aggregate knowledge from a set of pre-trained expert models to guide the training of a universal student model for a downstream complex task. 
Formally, each teacher specializes in its distinct task, $T_k$, defined as a dataset $\mathcal{D}_{k}$ with annotations $Y_k \in \mathcal{Y}_{k}$. 
The student model aims to effectively learn the joint task $\operatorname{T}=\left \{T_{1},\cdots, T_{k},\cdots, T_{K} \right \}$, 
defined as the dataset $\mathcal{D}=\bigcup_{k=1}^{K}\mathcal{D}_{k}$ and labels $\mathcal{Y}=\bigcup_{k=1}^{K}\mathcal{Y}_{k}$.
It is important to note that arbitrary dataset $\mathcal{D}_i$ and 
 $\mathcal{D}_j$ are mutually exclusive, i.e., $\mathcal{Y}_i \cap \mathcal{Y}_j = \oslash$.
 This process can be accomplished by optimizing the following function,
\begin{equation}
\mathcal{L}=\mathbb{E}_{\substack{X\sim P(X)\\ Y\sim P(Y)}}\left[\operatorname{CE}\left(g\left(f\left(X\right)\right), Y\right)+\sum_{k=1}^{K}\mathcal{L}_{align}\left(f(X), f^{T_k}(X) \right) \right] + 
\lambda \cdot \operatorname{Reg}(\Theta),
\end{equation}
where $X\in\mathcal{D}$ and $Y \in \mathcal{Y}$ are input and the ground truth, respectively; $f, f^{T_k}, g$ denote the student feature extractor, the $k$-th teacher feature extractor, and the predictor respectively; $\operatorname{CE}$ represents the Cross-Entropy used in classification or segmentation task; $\mathcal{L}_{align}$ is a feature alignment term that enables the student to learn discriminative knowledge from its teachers in a shared representative space; and a regularization term $\operatorname{Reg}$ is added to penalize abrupt changes in the neural network weights, $\Theta$. Following the method in~\cite{gao2024collaborative}, KL-divergence is used as a regularization term.
\begin{figure*}[t]
\captionsetup[subfigure]{position=above,justification=raggedright,singlelinecheck=off}
    \centering
    \subcaptionbox{\textbf{Stage 1}: nnUNet-based pre-training}[0.97\linewidth]{
        \includegraphics[width=\linewidth]{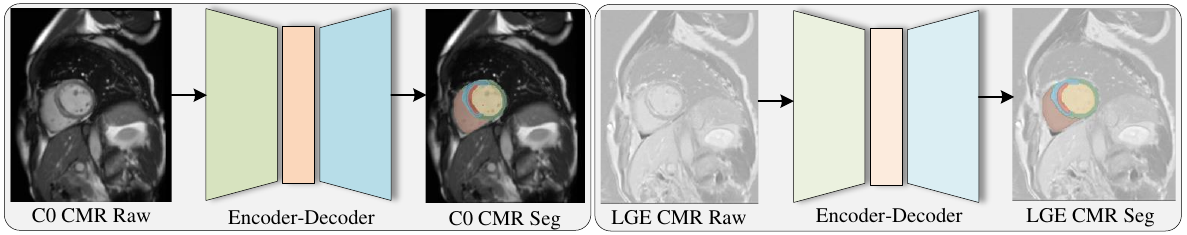} 
        \label{framework_petrain}
    }
    \subcaptionbox{\textbf{Stage 2}: Knowledge Adaptive amalgamation of Experts}[0.97\linewidth]{
        \includegraphics[width=\linewidth]{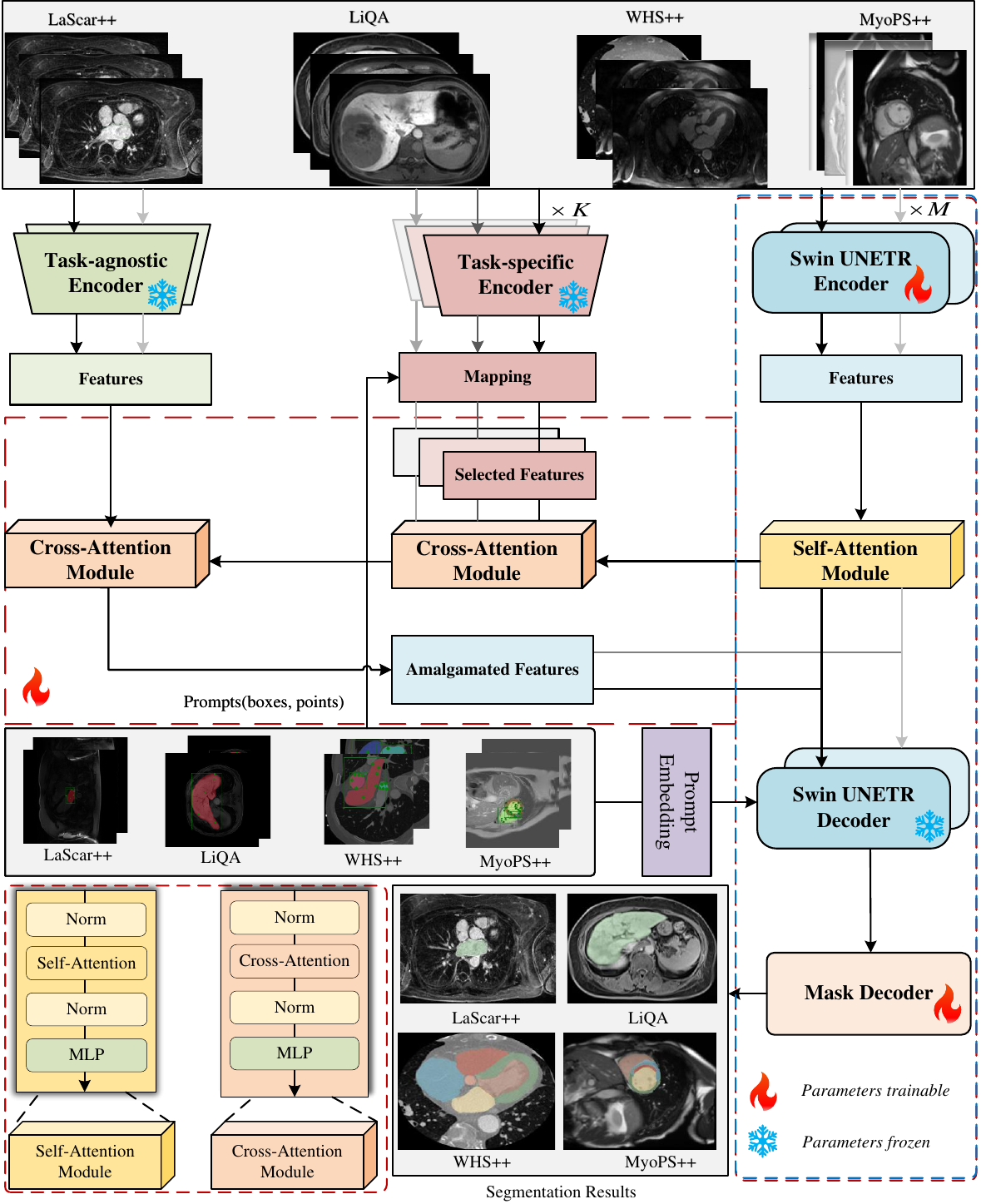} 
        \label{framework_train_and_infer}
    }
    \caption{(a) Input sub-task datasets (e.g., MyoPS$++$ split by modality, sequence) into the nnUNet framework, train them to obtain task-specific 
    expert models via cross-entropy loss. 
    (b) The knowledge adaptive amalgamation of experts framework includes a training phase (red dashed rectangular) and inference phase 
    (blue dashed rectangular). In the training phase, the target student network model adaptively merges task-specific and task-agnostic expert knowledge 
    based on the hierarchical attention module. In the inference phase, only a single target model is used for inference of input medical images.}
    \label{fig1:all}
\end{figure*}
Compared to a task-specific expert model, the student model exhibits greater robustness to domain shifts 
by aggregating the domain knowledge of all experts (e.g., aggregating the single-modal knowledge of each expert model in a multi-modal task) 
and effectively mitigates the adverse effects of heterogeneity or task differences among pre-trained experts.
\subsection{Adaptive Amalgamation of Pre-trained Experts}
In this work, we amalgamate the knowledge of the foundation model and multi-expert models for real-world medical image segmentation tasks. 
The overall workflow of our framework is shown in Figure. \ref{fig1:all}, which includes two stages: nnUNet-based pre-training and knowledge adaptive amalgamation of experts.
Specifically, we first train one nnUNet for each sub-task in \textbf{Stage 1} to obtain the task-specific experts, $\lbrace \mathcal{E}^{T_k}\rbrace_{k=1}^K$. Then, given a batch of images, $\lbrace X^{T_1} \in \mathcal D_k\rbrace_{k=1}^K$, we use the corresponding expert models to extract task-specific expert features, $f_e=\lbrace \mathcal{E}^{T_k}(X^{T_k})\rbrace_{k=1}^K$. To sufficiently utilize expert knowledge, we introduce the adaptive amalgamation to integrate the knowledge of experts into the target foundation model in \textbf{Stage 2}. The adaptive knowledge amalgamation is achieved through three hierarchical attention mechanisms, as shown in Fig.~\ref{fig1:all}(b). First, we perform self-attention, $\mathcal{A}$, on the features from the encoder, $\mathcal S_{enc}$, of the foundation model to get the important features, $f_s=\lbrace \mathcal{A}\circ\mathcal{S}_{enc}(X^{T_k})\rbrace_{k=1}^K$. Then these features are put into a cross-attention module as a query of all experts $f_e$. The cross-attention module is designed for the foundation model to obtain expert knowledge of specific segmentation tasks. Finally, we put these features into a cross-attention module as a query of an external expert to obtain the final merged feature $f_m$. The external expert is a medical image segmentation model and we use the total segmentator here \cite{d2024totalsegmentator}. With the second cross-attention module, the features further obtain the important features for general medical image segmentation. To make the foundation model learn the knowledge from both external and task-specific experts, we introduce an align loss measuring the distance between the feature of foundation models $f_s$ and the merged feature $f_m$ for segmentation as follows:
\begin{equation}
    \mathcal{L}_{align} = \| f_s - f_m \|.
\end{equation}

Furthermore, merged features are fed into the Swin UNETR decoder layers, where they are added and normed with the pre-encoded prompts before reaching the final segmentation layer. 
These prompts are generated by the ground-truth of medical image segmentation and encoded, with the weights of the encoding layer derived from MedSAM pre-training. 
Subsequently, the encoded prompt, along with the fused features, is passed through a lightweight, learnable mask decoder to produce the final segmentation output. 
The training workflow, including this process, is depicted in the red dashed rectangular box in Figure. \ref{fig1:all}.
\subsection{Universal Inference}
As the blue dashed rectangular shown in Figure. \ref{fig1:all}, during the inference phase of the foundation model, the knowledge amalgamation blocks containing cross-attention are entirely omitted. 
Input images are processed solely through the Swin UNETR encoder, $\mathcal{S}_{enc}$, self-attention mechanisms, $\mathcal{A}$, and the Swin UNETR decoder, $\mathcal{S}_{dec}$, for feature representation. 
After that, the resulting features and prompt embeddings, $E_{pro}$, are fed simultaneously into the mask decoder, $\mathcal{M}_{dec}$, to produce the final segmentation results. Overall, given an arbitrary image, $X\in \mathcal D$, we have a universal segmentation, $Y=\mathcal{M}_{dec}(\mathcal{S}_{dec}\circ \mathcal{A}\circ \mathcal{S}_{enc}(X), E_{pro}) $, during inference stage.
\section{Experiments}
\subsection{Dataset and Implementation}
\subsubsection{Dataset} We evaluate our method on four datasets, i.e., \textbf{MyoPS$++$}, \textbf{WHS$++$}, \textbf{LiQA} and \textbf{LAScarQS$++$}. 
The detailed statistics are summarized as follows.

\textbf{MyoPS$++$} contains multi-sequence CMR images of 250 patients from 7 centers, denoted by A-G, respectively.
145 patients in this dataset had three CMR sequences, LGE, C0, and T2, 24 patients had two sequences, bSSFP and LGE, and the remaining 81 patients had only one LGE sequence.
Data and annotations from 6 centers, and data from the remaining center with no annotations, were equally divided into the validation and test sets.
\textbf{WHS$++$} includes 206 multi-modality whole heart images from 6 centers, including 104 CT and 102 MRI. 
The training set contains 40 CT and 46 MRI images, the validation set contains 30 CT and 20 MRI images, and the test set contains 34 CT and 36 MRI images.
\textbf{LiQA} was composed of 440 patients diagnosed with liver fibrosis who underwent multi-phase MRI scans. 
The training set contains 30 annotated images and 250 unannotated images, and the validation set contains 30 annotated images, along with 160 test cases, 120 of which are from the vendors provided in the training set, and 40 of which are from a third, unseen vendor, which are tested for model generalizability.
\textbf{LAScarQS$++$} includes 194 multicenter LGE MRI images with different sources denoted by A, B, and C, respectively.
The dataset consists of two tasks, one is to segment the LA cavity and the scar region, and another is to segment only the LA cavity. 
Specifically, the training set, validation set, and test set for task 1 are all from center A, with quantities of 60, 10, and 10, respectively.
The training set for task 2 is from center A with the number of 130, and the validation set consists of 10 LGE MRIs from center A and 10 LGE MRIs from center C: The test set has 14 LGE MRI images from center A, 20 LGE MRI images from center B, and 10 LGE MRI images from center C composed. 
It is worth stating that these datasets are very challenging due to the unaligned data from different centers or the presence of missing modalities or missing labels.
\subsubsection{Evaluation Metrics}
We mainly employ \textbf{Dice similarity (DSC)} and \textbf{Hausdorff distance (HD)} to evaluate the segmentation performance. 

\subsubsection{Implementation Details}
In the pre-training phase, we trained nnUNet models for each task~\cite{isensee2021nnu}, including \textit{2d}, \textit{3d\_fullres}, and \textit{3d\_lowres} architectures, which were used as task-specific experts. Specifically, for the MyoPS$++$ task, where sub-datasets from different centers have inconsistent labels and modalities, we divided it into three sub-datasets—A, B+C, and E+F+G—based on the label categories of the training dataset. The tasks were then partitioned for each sub-dataset, with 80\% selected as the training set and 20\% as the internal validation (InVal) set.
For the other tasks, the datasets were categorized into different sub-tasks based on the task or data modalities, resulting in a total of 10 pre-trained models. 
All data processing and training processes followed the specifications of nnUNet. 
Each pre-trained model was trained on a compute cluster using NVIDIA™ 4090 24GB GPUs. 
The training epochs were set to 1000, with a learning rate of 0.01, and the batch size was dynamically adjusted according to the split subtasks.
To ensure the model's ability to generalize to agnostic knowledge, we downloaded and executed the nnUNet \textit{3d\_fullres} models from Totalsegmentator~\cite{d2024totalsegmentator} as external expert models.

To train the Swin UNETR-based foundation model, we adhered to the workflow illustrated in Figure \ref{fig1:all}. 
All datasets were transformed using operations such as RandomCropByPosNeg, Spacingd, and RandRotated before being fed into the Swin UNETR Encoder for backbone feature extraction. 
The training process utilized a learning rate of $1 \times 10^{-4}$ with cosine decay, an embedding dimension of 48, the Adam optimizer, and a batch size of 8. 
The model was trained for 300 epochs on a computing cluster equipped with NVIDIA™ A800-SXM4-80GB GPUs. 
Hyperparameters included a region of interest (ROI) dimension set to $96 \times 96 \times D$, where $D$ is an integer multiple of 32. 
Notably, the orientation parameter was not set to PLS during training, as the orientation parameter of the MyoPS$++$ dataset was uncalibrated. 
Unless otherwise specified, parameters such as spacing and orientation were consistent with those used in the pre-trained nnUNet model. 
The framework was implemented using PyTorch and MONAI~\cite{cardoso2022monai}.

During the inference phase, the hierarchical attention module and all expert layers were omitted, with all parameters of the foundation model kept frozen. Only raw medical images and their corresponding prompts were required to generate task-specific prediction results. The predicted outcomes underwent necessary post-processing, including the maximization of connected regions, before being submitted to the Challenge public platform for external validation (ExVal) to obtain the final evaluation results.
\subsection{Comprehensive Multi-task Segmentation}
\begin{table}[t]
    \centering
    \small
\caption{Peformance comparison of multi-task segmentation across models. The results are reported in terms of Dice similarity coefficient (\%) on the internal validation (InVal) and external validation (ExVal) sets. “G” denotes generalization performance. “*” indicates external validation failure.} 
    \label{tab:comparison}
\begin{tabular}{llcccccc}
\toprule
\rowcolor{white} \textbf{Dataset} & \textbf{Task} & \multicolumn{2}{c}{\textbf{nnUNet}} & \multicolumn{2}{c}{\textbf{Swin UNETR}} & \multicolumn{2}{c}{\textbf{KA$^2$ER (Our)}} \\
\cline{3-4} \cline{5-6} \cline{7-8} 
 & & InVal & ExVal (G) & InVal & ExVal (G) & InVal & ExVal (G) \\
\midrule
\multirow{3}{*}{WHS$++$} & CT & 93.1 & 86.1 ($\downarrow$7.0) & 92.8 & 86.4 ($\downarrow$6.4) & 88.6 & 88.1 ($\downarrow$0.5) \\
 & MR & 87.6 & 88.7 ($\uparrow$1.1) & 82.4 & 80.6 ($\downarrow$1.8) & 88.5 &  79.6 ($\downarrow$8.9)  \\
 &CT+MR & 87.4 & 87.1 ($\downarrow$0.3) & - & - & - & - \\
\hline
\multirow{3}{*}{MyoPS$++$}  & B+C & 68.4 & 56.8 ($\downarrow$11.6) & 81.8 & * & 82.2 & * \\
& A & 73.3 & - & 85.8 & - & 85.8 & - \\

 & E+F+G & 69.8 & - & 93.2 & - & 93.2 & - \\
\hline
\multirow{3}{*}{LiQA} & Vendor A &  87.2 &86.5 ($\downarrow$0.7) & 98.1 & 90.5 ($\downarrow$7.6) & 98.4 & 91.1 ($\downarrow$7.3)\\
 & Vendor B1 & 95.6 & 96.1 ($\uparrow$0.5) & 96.1 & 93.0 ($\downarrow$3.1) & 98.0 & 95.8 ($\downarrow$2.2) \\
 & Vendor B2 & 95.9 & 94.9 ($\downarrow$1.0) & 96.6 & 93.8 ($\downarrow$2.8) &  97.6 & 92.4 ($\downarrow$5.1)  \\
& A+B1+B2 & 91.2 & 92.5 ($\uparrow$1.3) & 95.6 & -  & 98.0 & 93.1 ($\downarrow$4.9) \\
\hline
\multirow{4}{*}{LaScarQS$++$}
 & Task1\_scar &  47.0 & 14.7 ($\downarrow$32.3)&  71.0 & 50.4 ($\downarrow$20.6)& 52.8 & 51.1 ($\downarrow$1.7)\\
 & Task1\_cavity & 92.7 & - & 98.3 & - & 99.2 & - \\
 & Task2 & 92.5 & 87.5 ($\downarrow$5.0) & 97.6 & 87.6 ($\downarrow$10.0)& 92.6 & 87.5 ($\downarrow$5.1)\\
\bottomrule
\end{tabular}
\end{table}
We conducted experiments on four MICCAI'2024 CARE Challenge datasets to evaluate the KA$^2$ER framework's performance. Table \ref{tab:comparison} summarizes the results, comparing our model with nnUNet and Swin UNETR on the internal (InVal) and external (ExVal) validation sets, where "G" indicates generalization capability.
In the internal validation on WHS$++$, nnUNet achieved the best performance, with Dice scores of 93.1\% for CT and 87.6\% for MR. In contrast, our model achieved Dice scores of 88.6\% for CT and 88.5\% for MR, approximately 4\% lower in the CT modality. For other datasets, particularly in the MyoPS++ and Lascar\_task1 challenges, Swin UNETR outperformed nnUNet with scores of 81.8\% vs. 68.4\%, and 71.0\% vs. 47.0\%, respectively, while our model achieved 82.8\% and 52.8\% in these tasks. During external validation, both our model and Swin UNETR failed on the MyoPS$++$ dataset. This can be attributed to the inherent difficulty of distinguishing scar tissue from edema, as well as semantic conflicts arising from our model's use of bounding box prompts, where scar tissue is closely associated with edema. Furthermore, the combination of internal and external validation suggests that while nnUNet performs well in specific tasks, its generalization ability is limited.
Our proposed K$^2$AE model, integrating knowledge-based and prompt-based techniques, outperforms the baseline in most tasks and demonstrates strong competitiveness.

\begin{figure}[t!]
    \centering
\includegraphics[width=0.99\textwidth]{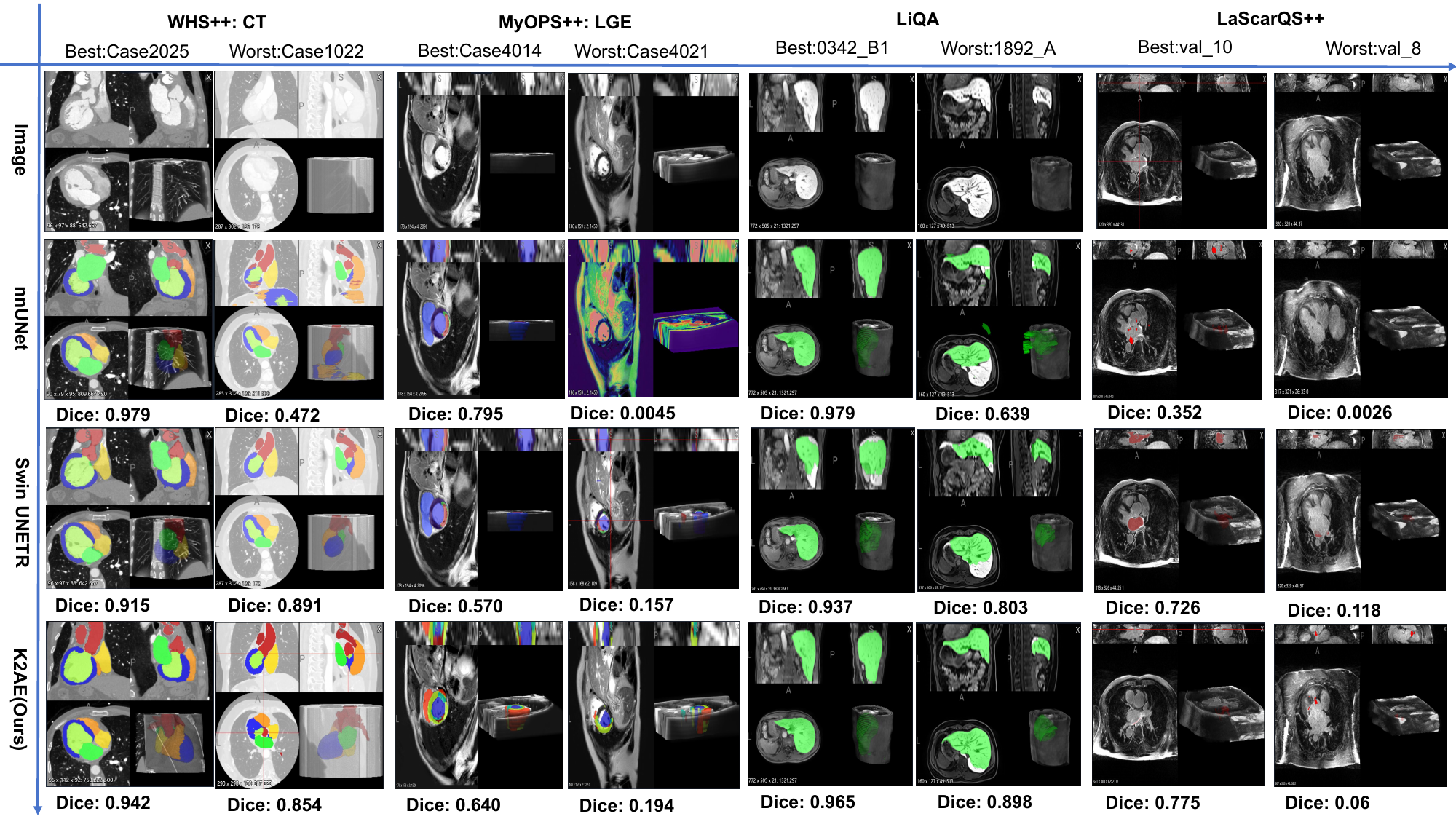}
    \caption{Visualization of comprehensive multi-task segmentation results. The results are reported in terms of DSC for best and worst cases.}
    \label{fig:visual_segmentations}
\end{figure}
Figure \ref{fig:visual_segmentations} visualizes best and worst case comparisons for WHS$++$, MyoPS$++$, LiQA, and LaScarQS$++$ datasets, using the nnUNet model's segmentation results on the ExVal validation set as a baseline. 
Specifically, the first row of Figure \ref{fig:visual_segmentations} represents original images, while the 2-4 rows show the segmentation results of different models. 
nnUNet exhibited significant variability, especially in MyoPS$++$ and Lascar\_task1, where the worst-case Dice scores were 0.45\% and 0.26\%. Swin UNETR showed greater robustness against data perturbations, which can be attributed to its self-supervised pretraining on large-scale datasets.
In contrast, our model produced consistently better results, with stronger generalization capability, aligning with the experimental results shown in Table \ref{tab:comparison}.

\subsection{Challenging multi-structure segmentation}

We also evaluate the performance of the proposed method on the challenging multi-structure segmentation tasks.
On the WHS$++$ dataset, models are trained to segment 7 targets, including the Left Ventricular Blood Cavity (LV), Right Ventricular Blood Cavity (RV), Left Atrial Blood Cavity (LA),
Right Atrial Blood Cavity (RA), Myocardium of the Left Ventricle (Myo), Ascending Aorta (AO), and Pulmonary Artery (PA). 
As shown in Table.~\ref{tab:whs_comparison}, nnUNet and SwinUNETR perform better than our KA$^2$ER on the CT modality, while KA$^2$ER outperforms SwinUNETR and is compariable with nnUnet on the MR modality.
It is worth noting that the KA$^2$ER model is an universal model that can be applied to not only different 
modalities of both CT and MR but also different tasks, while the nnUNet and SwinUNETR models are task-specific models. 
On the MyoPS$++$ dataset, as shown in Table.~\ref{tab:myops_comparison}, our model outperforms the nnUNet and SwinUNETR models. Specifically, for the Edema segemntation task, KA$^2$ER achieves a Dice score of 75.8\%, which is 43.5\% higher than the nnUNet. 

\begin{table}[t!]
\small
    \centering
    \caption{Performance comparison of different models on InVal of MyoPS++. 
     The results are reported in terms of DSC (\%) and HD (mm).}
    \label{tab:myops_comparison}
    \resizebox{\linewidth}{!}{
    \begin{tabular}{cccccccccccccc}
        \toprule
        \rowcolor{white} \textbf{Center} & \textbf{Model} &\multicolumn{2}{c}{\textbf{LV}}  &\multicolumn{2}{c}{\textbf{Scar}}
        &\multicolumn{2}{c}{\textbf{Myo}}
        &\multicolumn{2}{c}{\textbf{RV}}
        &\multicolumn{2}{c}{\textbf{Edema}}
        &\multicolumn{2}{c}{\textbf{AVG}}\\
        \cline{3-4} \cline{5-6} \cline{7-8} \cline{9-10} \cline{11-12}
        & & DSC & HD & DSC & HD & DSC & HD & DSC & HD & DSC & HD & DSC & HD \\
        \midrule
        \multirow{3}{*}{B + C} & nnUNet & 89.6 & 8.46 & 64.8 & 50.2 & 73.3 & 23.64 & 82.0 & 13.92 & 32.3 & 125.2 & 68.4 & 44.2 \\
         & Swin UNETR &  &  &  &  &  &  &  &  &  &  & 81.8 & 9.21\\
         & KA$^2$ER & 88.8 & - & 79.4 & - & 80.1 & - & 87.1 & - & 75.8 & - & 82.2 & 8.87 \\
        \bottomrule
    \end{tabular}}
\end{table}
\begin{table}[t!]
    \centering
    \small
    \vspace{-2em}
    \caption{Performance comparison of different models on InVal of WHS++. The results are reported in terms of DSC (\%) and HD (mm).}
    \label{tab:whs_comparison}
    \resizebox{\linewidth}{!}{
    \begin{tabular}{ccccccccccccccccccccc}
        \toprule
        \rowcolor{white} \textbf{Modality} & \textbf{Model} & \multicolumn{2}{c}{\textbf{LV}} & \multicolumn{2}{c}{\textbf{RV}} & \multicolumn{2}{c}{\textbf{LA}} & \multicolumn{2}{c}{\textbf{RA}} & \multicolumn{2}{c}{\textbf{Myo}} & \multicolumn{2}{c}{\textbf{AO}} & \multicolumn{2}{c}{\textbf{PA}} & \multicolumn{2}{c}{\textbf{AVG}}\\
        \cline{3-4} \cline{5-6} \cline{7-8} \cline{9-10} \cline{11-12} \cline{13-14} \cline{15-16} 
        & & DSC & HD & DSC & HD & DSC & HD & DSC & HD & DSC & HD & DSC & HD & DSC & HD & DSC & HD
        \\
        \midrule
        \multirow{3}{*}{CT} & nnUnet & 93.0 & 7.96 & 85.2 & 13.16 & 93.3 & 13.69 & 85.8 & 16.78 & 82.2 & 19.59 & 96.1 & 9.25 & 84.2 & 15.56 & 89.9 & 13.71 \\
        & Swin UNETR &  &  &  &  &  &  &  &  &  &  &  &  &  &  & 92.8 & 13.16 \\
        & KA$^2$ER & 86.3 & 8.14 & 88.4 & 12.57 & 89.1 & 15.78 & 88.7 & 15.25 & 87.5 & 16.28 & 93.0 & 11.61 & 87.2 & 14.93 & 88.6 & 13.50 \\
        \cline{2-18}
        \multirow{3}{*}{MR} & nnUnet & 95.0 & 7.01 & 91.4 & 13.82 & 91.4 & 11.58 & 91.6 & 14.36 & 87.8 & 15.27 & 84.3 & 5.82 & 83.1 & 7.24 & 89.2 & 10.72 \\
        & Swin UNETR &  &  &  &  &  &  &   &  &  &  &  &  &  &  & 82.4 & 21.55 \\
        & KA$^2$ER & 90.4 & 8.82 & 87.3 & 17.21 & 89.3 & 14.56 & 88.2 & 15.11 & 84.3 & 17.29 & 89.9 & 3.55 & 90.3 & 4.68 & 88.5 & 11.61 \\
        \bottomrule
    \end{tabular}}

\end{table}
\subsubsection{\ackname} This research was partially supported by National Natural Science Foundation of China under grants No. 62176231 and No. 82202984, Zhejiang Key R\&D Program of China under grant No. 2023C03053 and No. 2024SSYS0026.
\section{Conclusion}
In this work, we propose a novel knowledge adaptive amalgamation of experts (KA$^2$ER) framework for comprehensive multi-task medical image segmentation.
The proposed framework consists of two stages: nnUNet-based pre-training and knowledge adaptive amalgamation of experts.
The pre-training stage involves training task-specific expert models using the nnUNet framework, while the knowledge adaptive amalgamation of experts stage involves integrating the knowledge of task-specific and task-agnostic experts into a target foundation model.
The proposed framework is evaluated on four datasets from the MICCAI'2024 CARE challenge, demonstrating competitive performance across various datasets.
The results show that the proposed KA$^2$ER framework outperforms the baseline models on most tasks, achieving state-of-the-art performance in comprehensive multi-task medical image segmentation.

\clearpage

%
\bibliographystyle{splncs04}
\bibliography{ka2er}

\begin{thebibliography}{10}
\providecommand{\url}[1]{\texttt{#1}}
\providecommand{\urlprefix}{URL }
\providecommand{\doi}[1]{https://doi.org/#1}

\bibitem{boyle2019computationally}
Boyle, P.M., Zghaib, T., Zahid, S., Ali, R.L., Deng, D., Franceschi, W.H., Hakim, J.B., Murphy, M.J., Prakosa, A., Zimmerman, S.L., et~al.: Computationally guided personalized targeted ablation of persistent atrial fibrillation. Nature biomedical engineering  \textbf{3}(11),  870--879 (2019)

\bibitem{cao2023large}
Cao, K., Xia, Y., Yao, J., Han, X., Lambert, L., Zhang, T., Tang, W., Jin, G., Jiang, H., Fang, X., et~al.: Large-scale pancreatic cancer detection via non-contrast ct and deep learning. Nature medicine  \textbf{29}(12),  3033--3043 (2023)

\bibitem{cardoso2022monai}
Cardoso, M.J., Li, W., Brown, R., Ma, N., Kerfoot, E., Wang, Y., Murrey, B., Myronenko, A., Zhao, C., Yang, D., et~al.: Monai: An open-source framework for deep learning in healthcare. arXiv preprint arXiv:2211.02701  (2022)

\bibitem{d2024totalsegmentator}
D'Antonoli, T.A., Berger, L.K., Indrakanti, A.K., Vishwanathan, N., Wei{\ss}, J., Jung, M., Berkarda, Z., Rau, A., Reisert, M., K{\"u}stner, T., et~al.: Totalsegmentator mri: Sequence-independent segmentation of 59 anatomical structures in mr images. arXiv preprint arXiv:2405.19492  (2024)

\bibitem{du2023segvol}
Du, Y., Bai, F., Huang, T., Zhao, B.: Segvol: Universal and interactive volumetric medical image segmentation. arXiv preprint arXiv:2311.13385  (2023)

\bibitem{gao2024collaborative}
Gao, S., Fu, Y., Liu, K., Gao, W., Xu, H., Wu, J., Han, Y.: Collaborative knowledge amalgamation: Preserving discriminability and transferability in unsupervised learning. Information Sciences  \textbf{669},  120564 (2024)

\bibitem{gao2023contrastive}
Gao, S., Fu, Y., Liu, K., Han, Y.: Contrastive knowledge amalgamation for unsupervised image classification. In: International Conference on Artificial Neural Networks. pp. 192--204. Springer (2023)

\bibitem{gao2022joint}
Gao, S., Zhou, H., Gao, Y., Zhuang, X.: Joint modeling of image and label statistics for enhancing model generalizability of medical image segmentation. In: International Conference on Medical Image Computing and Computer-Assisted Intervention. pp. 360--369. Springer (2022)

\bibitem{GAO2023BayeSeg}
Gao, S., Zhou, H., Gao, Y., Zhuang, X.: Bayeseg: Bayesian modeling for medical image segmentation with interpretable generalizability. Medical Image Analysis  \textbf{89},  102889 (2023)

\bibitem{HintonVD15}
Hinton, G., Vinyals, O., Dean, J.: Distilling the knowledge in a neural network. arXiv preprint arXiv:1503.02531  (2015)

\bibitem{isensee2021nnu}
Isensee, F., Jaeger, P.F., Kohl, S.A., Petersen, J., Maier-Hein, K.H.: nnu-net: a self-configuring method for deep learning-based biomedical image segmentation. Nature methods  \textbf{18}(2),  203--211 (2021)

\bibitem{kirillov2023segment}
Kirillov, A., Mintun, E., Ravi, N., Mao, H., Rolland, C., Gustafson, L., Xiao, T., Whitehead, S., Berg, A.C., Lo, W.Y., et~al.: Segment anything. In: Proceedings of the IEEE/CVF International Conference on Computer Vision. pp. 4015--4026 (2023)

\bibitem{li2021AtrialGeneral}
Li, L., Zimmer, V.A., Schnabel, J.A., Zhuang, X.: Atrialgeneral: domain generalization for left atrial segmentation of multi-center lge mris. In: Medical Image Computing and Computer Assisted Intervention--MICCAI 2021: 24th International Conference, Strasbourg, France, September 27--October 1, 2021, Proceedings, Part VI 24. pp. 557--566. Springer (2021)

\bibitem{li2022AtrialJSQnet}
Li, L., Zimmer, V.A., Schnabel, J.A., Zhuang, X.: Atrialjsqnet: a new framework for joint segmentation and quantification of left atrium and scars incorporating spatial and shape information. Medical image analysis  \textbf{76},  102303 (2022)

\bibitem{li2022Survey}
Li, L., Zimmer, V.A., Schnabel, J.A., Zhuang, X.: Medical image analysis on left atrial lge mri for atrial fibrillation studies: A review. Medical image analysis  \textbf{77},  102360 (2022)

\bibitem{liu2024merit}
Liu, Y., Gao, Z., Shi, N., Wu, F., Shi, Y., Chen, Q., Zhuang, X.: Merit: Multi-view evidential learning for reliable and interpretable liver fibrosis staging (2024)

\bibitem{ma2024segment}
Ma, J., He, Y., Li, F., Han, L., You, C., Wang, B.: Segment anything in medical images. Nature Communications  \textbf{15}(1), ~654 (2024)

\bibitem{qiu2023myops}
Qiu, J., Li, L., Wang, S., Zhang, K., Chen, Y., Yang, S., Zhuang, X.: Myops-net: Myocardial pathology segmentation with flexible combination of multi-sequence cmr images. Medical image analysis  \textbf{84},  102694 (2023)

\bibitem{ronneberger2015u}
Ronneberger, O., Fischer, P., Brox, T.: U-net: Convolutional networks for biomedical image segmentation. In: Medical image computing and computer-assisted intervention--MICCAI 2015: 18th international conference, Munich, Germany, October 5-9, 2015, proceedings, part III 18. pp. 234--241. Springer (2015)

\bibitem{shen2019amalgamating}
Shen, C., Wang, X., Song, J., Sun, L., Song, M.: Amalgamating knowledge towards comprehensive classification. In: Proceedings of the AAAI Conference on Artificial Intelligence. pp. 3068--3075. {AAAI} Press, Honolulu, Hawaii, USA (2019)

\bibitem{tang2022self}
Tang, Y., Yang, D., Li, W., Roth, H.R., Landman, B., Xu, D., Nath, V., Hatamizadeh, A.: Self-supervised pre-training of swin transformers for 3d medical image analysis. In: Proceedings of the IEEE/CVF conference on computer vision and pattern recognition. pp. 20730--20740 (2022)

\bibitem{Wu2023SemiSL}
Wu, F., Zhuang, X.: Minimizing estimated risks on unlabeled data: A new formulation for semi-supervised medical image segmentation. IEEE Transactions on Pattern Analysis and Machine Intelligence  \textbf{45}(5),  6021--6036 (2023)

\bibitem{xu2023group}
Xu, R., Yang, K., Liu, K., He, F.: Group equivariant vision transformer. In: Conference on Uncertainty in Artificial Intelligence (2023)

\bibitem{xu2024survey}
Xu, X., Li, M., Tao, C., Shen, T., Cheng, R., Li, J., Xu, C., Tao, D., Zhou, T.: A survey on knowledge distillation of large language models. arXiv preprint arXiv:2402.13116  (2024)

\bibitem{yang2023knowledge}
Yang, Z., Zeng, A., Li, Z., Zhang, T., Yuan, C., Li, Y.: From knowledge distillation to self-knowledge distillation: A unified approach with normalized loss and customized soft labels. In: Proceedings of the IEEE/CVF International Conference on Computer Vision. pp. 17185--17194 (2023)

\bibitem{zhang2023knowledge}
Zhang, H., Mao, F., Xue, M., Fang, G., Feng, Z., Song, J., Song, M.: Knowledge amalgamation for object detection with transformers. IEEE Transactions on Image Processing  \textbf{32},  2093--2106 (2023)

\bibitem{zhuang2013challenges}
Zhuang, X.: Challenges and methodologies of fully automatic whole heart segmentation: a review. Journal of healthcare engineering  \textbf{4}(3),  371--407 (2013)

\bibitem{zhuang2019multivariate}
Zhuang, X.: Multivariate mixture model for myocardial segmentation combining multi-source images. IEEE transactions on pattern analysis and machine intelligence  \textbf{41}(12),  2933--2946 (2019)

\bibitem{Zhuang2019evaluation}
Zhuang, X., Li, L., Payer, C., Stern, D., Urschler, M., Heinrich, M.P., Oster, J., Wang, C., Smedby, {\"O}., Bian, C., et~al.: Evaluation of algorithms for multi-modality whole heart segmentation: an open-access grand challenge. Medical image analysis  \textbf{58},  101537 (2019)

\end{thebibliography}

\end{document}